# Multi-fidelity Fourier Neural Operator for Fast Modeling of Large-Scale Geological Carbon Storage

Hewei Tang[1*], Qingkai Kong[1*] and Joseph P. Morris[1]

[1]Atmospheric, Earth, and Energy Division, Lawrence Livermore National Laboratory, Livermore, CA 94550, USA

\* Corresponding authors: tang39@llnl.gov  kong11@llnl.gov

## Abstract

Deep learning-based surrogate models have been widely applied in geological carbon storage (GCS) problems to accelerate the prediction of reservoir pressure and $CO_2$ plume migration. Large amounts of data from physics-based numerical simulators are required to train a model to accurately predict the complex physical behaviors associated with this process. In practice, the available training data are always limited in large-scale 3D problems due to the high computational cost. Therefore, we propose to use a multi-fidelity Fourier neural operator (FNO) to solve large-scale GCS problems with more affordable multi-fidelity training datasets. FNO has a desirable grid-invariant property, which simplifies the transfer learning procedure between datasets with different discretization. We first test the model efficacy on a GCS reservoir model being discretized into 110k grid cells. The multi-fidelity model can predict with accuracy comparable to a high-fidelity model trained with the same amount of high-fidelity data with 81% less data generation costs. We further test the generalizability of the multi-fidelity model on a same reservoir model with a finer discretization of 1 million grid cells. This case was made more challenging by employing high-fidelity and low-fidelity datasets generated by different geostatistical models and reservoir simulators. We observe that the multi-fidelity FNO model can predict pressure fields with reasonable accuracy even when the high-fidelity data are extremely limited. The findings of this study can help for better understanding of the transferability of multi-fidelity deep learning surrogate models.

1. Introduction

Geological carbon storage (GCS) plays an important role in mitigating global climate change (Zhou and Birkholzer, 2011). Numerical modeling of GCS process is crucial in the permitting phase as well as the subsequent reservoir monitoring and management process. In a physics-based numerical simulator, the GCS process is modeled as multiphase flow in a subsurface porous media governed by partial differential equations (PDEs). High-resolution spatial discretization in three dimensions (3D) is required to solve the nonlinear governing equations. The computational cost of a large 3D GCS problem is usually high and is computational prohibitive to run within inverse or optimization problems, which requires thousands of runs.

Deep learning-based surrogate models can provide significant speedup in predicting state variables (such as pressure and $CO_2$ saturation) compared to traditional numerical simulators. Convolutional neural networks (CNNs) have been successfully applied to predict fluid flow behavior in subsurface porous media. The high-dimensional surrogate modeling problem is viewed as an image-to-image regression task (Zhu and Zabaras, 2018). In the context of GCS applications, Mo et al. (2019) proposed a convolutional encoder-decoder framework to predict fluid pressure and $CO_2$ saturation in a two-dimensional open flow domain. Tang et al. (2022) applied a recurrent residual U-Net model to predict $CO_2$ plume migration, pressure distribution and displacement in a 3D reservoir domain. All these CNN-based models are grid dependent, which requiring the network to be redesigned and retrained whenever the spatial discretization of the numerical model changes.

Neural operator-based approaches such as Fourier Neural Operator (FNO) methods possess the desirable grid-invariant property (Li et al., 2021). They have also been successfully applied in GCS modeling. Yan et al. (2022) applied a 2D FNO structure to predict layer-wise pressure and $CO_2$ saturation distribution in a 3D reservoir on a series of independent 2D horizontal slices. Witte et al. (2023) applied a 4D Wavelet Neural Operator to predict spatial-temporal $CO_2$ saturation around a single injector in a 3D reservoir domain with 230,000 grid cells. It took them 18k CPU hours to generate 4000 training samples. Wen et al. (2023) proposed a nested 4D FNO structure to predict high-resolution 3D pressure and $CO_2$ saturation fields considering multiple injectors. Different levels of local grid refinement were considered with ~8000 simulations generated for the training purposes.

Both CNN-based and neural operator-based models rely on large amounts of data to train a model to achieve sufficient accuracy. However, for a large-scale GCS simulation with more than 1 million grid cells, it is computational prohibitive to generate thousands of realizations. Multi-fidelity modeling approaches which utilize a large set of low-fidelity data, and a small number of high-fidelity data can potentially reduce the cost of training data acquisition. Here, high-fidelity and low-fidelity data are specifically referred to simulation results acquired based on fine and coarse discretization. Song and Tartakovsky (2022) proposed a transfer learning approach to train different parts of a CNN-based surrogate model for 2D multiphase flow based on a multi-fidelity dataset. The approach contains three training phases and introduces an additional temporary layer to accommodate the structural difference between the low-fidelity model and the high-fidelity model. Jiang and Durlofsky (2023) applied a similar three-step transfer learning strategy to model 3D two-phase flow problems based on a recurrent residual U-Net model. The additional complexity of the temporary layer and training steps required in the above multi-fidelity models can be simplified by applying the FNO as the base model. This is due to its grid-invariant property, which enables the low-fidelity and high-fidelity data to utilize the same network structure without structural modifications. The idea of training FNO on multi-fidelity datasets has been applied to several engineering applications such as temperature field prediction, airfoil flows, and laminar single-cylinder wake (Lyu et al., 2023). They concluded that both increasing the quality of low-fidelity data and the amount of high-fidelity training data can help improve the accuracy of multi-fidelity FNO models.

In this work, we explore the feasibility of applying a multi-fidelity FNO model to solve large-scale GCS problems. Compared to other engineering applications, subsurface flow models need to additionally consider heterogeneity within the spatial domain (e.g., porosity and permeability). Different grid resolutions of the reservoir domain can result in different levels of spatial heterogeneity and thus different numerical simulation results. In this sense, a low-fidelity dataset is not necessarily a 'low-resolution' representation of the high-fidelity dataset. They are effectively solutions to different problems, making multi-fidelity training more challenging. To obtain a low-fidelity dataset as similar as possible to the high-fidelity dataset, a common practice is to directly upscale the realizations of random input fields at fine scale. For example, Jiang and Durlofsky (2023) applied a global transmissibility upscaling procedure to directly generate low-fidelity geomodels from high-fidelity geomodels.

However, having access to thousands of high-fidelity geomodel realizations is also challenging in a large-scale GCS problem. Geomodels are usually generated from objective-based models or variogram-based models such as sequential gaussian simulation (SGS). The computational cost associated with geomodel generation increases sharply as the number of grid cells increases. Specifically, the computational cost to generate a realization with *N* grid cells is about O($N^3$) using SGS (Bai and Tahmasebi, 2022). Taking this constraint into account, we further explore the feasibility of utilizing low-fidelity geomodels directly generated from geostatistical models rather than upscaled from high-fidelity geomodeles. This is a more challenging case since the two datasets are not directly correlated.

The purpose of this paper is to build a multi-fidelity FNO surrogate model for large-scale GCS applications. Section 2 introduces the governing equations of GCS processes, the FNO model structure, and the multi-fidelity training framework. Section 3 introduces a specific numerical example to test the efficacy of the multi-fidelity model. The model performance and computational costs are summarized in Section 4. Section 5 discusses the generalizability of the multi-fidelity training framework to indirectly correlated datasets. In section 6, we conclude the work with a few remarks.

## 2. Methodology

### 2.1 Governing Equations of GCS Process

Injecting $CO_2$ into deep saline aquifer can be modeled as a two-phase flow problem in porous media. We consider an aqueous phase containing both $H_2O$ and $CO_2$ components and a non-aqueous $CO_2$ phase containing only $CO_2$ component. The $CO_2$ phase can be in supercritical, liquid or gas state. Note that in the actual physical process the non-aqueous phase can contain $H_2O$ as well, especially in the near-well region, but $H_2O$ evaporation is not considered in our models. The $CO_2$ solubility in brine is calculated using the model of Duan and Sun (2023). The mass conservation equation for component *c* (*c* can be either $CO_2$ or $H_2O$) is:

$$\frac{\partial \phi}{\partial t}\left(\sum_l \rho_l m_{cl} S_l\right) + \nabla \cdot \left(\sum_l \rho_l m_{cl} \mathbf{v}_l\right) - \sum_l \rho_l m_{cl} q_l = 0, \tag{1}$$

where $m_{cl}$ denotes mass fraction of component $c$ in phase $l$, $S_l$ is saturation of phase $l$, $\phi$ is porosity, and $q_l$ is injected phase volume. The phase transport velocity in porous media $\mathbf{v}_l$ is usually approximated by Darcy's law:

$$\mathbf{v}_l = -\frac{k k_{rl}}{\mu_l}(\nabla p_l - \rho_l g \nabla z), \tag{2}$$

where $p_l$ is the pressure of phase $l$, $\rho_l$ is the phase density, $\mu_l$ is the phase viscosity, $k$ is the absolute permeability of rock, $k_{rl}$ is the relative permeability of phase $l$, and z is depth. The subsurface porous media are usually highly heterogeneous, which means that $\phi$ and $k$ can vary significantly spatially. The pressure of the two phases can be correlated through capillary pressure ($p_c$), and the sum of the phase saturations is one.

To obtain a spatial distribution of pressure and $CO_2$ saturation with desired resolution, the model domain is discretized with finite volume methods to form a system of discrete non-linear equations. Newton-based methods that linearize the equations are typically applied to solve the system of non-linear equations iteratively. As discretization becomes finer, the computational cost of linear solvers increases sharply, especially for 3D problems. Therefore, we intend to build surrogate models to approximate the non-linear mapping between input parameters ($\phi$, $k$, and injection schedule) and the solution space (pressure and $CO_2$ saturation) for large-scale 3D problems with fine discretization. We will develop two separate surrogate models for pressure and $CO_2$ saturation predictions.

## 2.2 Fourier Neural Operator

The FNO belongs to the family of neural operators, where the deep neural network learns the solution operator of a given PDE through the finite collection of input-output pairs, *a(x)*, and *u(x)* (Li et al., 2021). Here, *x* represents the spatial discretization of domain *D*. A neural operator is formulated as an iterative architecture $v_0 \to v_1 \to \cdots \to v_T \to v_{T+1}$. Here $v_0$ is usually defined using a linear transformer *P*:

$$v_0(x) = P(a(x)), \tag{3}$$

where the input space *a(x)* is lifted by a fully connected neural network to a higher dimension. Here, $v_t$ ($t = 1, 2, \ldots T$) is a neural operator layer defined as:

$$v_{t+1}(x) = \sigma(W v_t(x) + (\mathcal{K} v_t)(x)), \tag{4}$$

where *W* is a linear operator, $\mathcal{K}$ is a kernel integral operator, and $\sigma$ is a non-linear activation function. Using the FNO, the kernel integral operator is replaced by an operator defined in Fourier space (Li et al., 2021; Zhao et al., 2023):

$$(\mathcal{K} v_t)(x) = \mathcal{F}^{-1}(R \cdot T_K(\mathcal{F} v_t))(x), \tag{5}$$

where $\mathcal{F}$ denotes the Fourier transform operator, and $T_K$ is a truncation operation to filter out Fourier modes higher than *K*. *R* is a learnable transformation in Fourier space, and $v_{t+1}$ is a transformation to project $v_t$ onto the desired output dimension by applying a linear operator *Q*.

$$v_{t+1}(x) = Q(v_t(x)), \tag{6}$$

The FNO has desirable grid-invariant features, which means that the model can be evaluated at any query points that are not necessarily members of the set of training grids. This feature allows the model to be easily transferred between different discretization.

## 2.3 Multi-fidelity Training Framework

The multi-fidelity training framework of FNO is shown in Figure 1. The framework contains two steps: one is pre-training, and the other is fine-tuning. The first step of pre-training is based on a low-fidelity dataset containing a large number of low fidelity data pairs that are generated based on a coarse discretization of the computational domain ($D^{LF} = \{(a(x_i^{LF}), u(x_i^{LF}))\}_{i=1}^{N}$). The low fidelity model is pre-trained based on relative $l_p$-loss defined as:

$$L\left(u(x_i^{LF}), \hat{u}(x_i^{LF})\right) = \frac{\left\|u(x_i^{LF}) - \hat{u}(x_i^{LF})\right\|_p}{\left\|u(x_i^{LF})\right\|_p} \quad (7)$$

The second step of fine-tuning involves a small number of high-fidelity data pairs that are generated based on a refined discretization of the computation domain ($D^{HF} = \{(a(x_i^{HF}), u(x_i^{HF}))\}_{i=1}^{N}$). The pre-trained low-fidelity model is loaded as an initial model for fine-tuning. The multi-fidelity model is obtained based on relative $l_p$-loss defined as:

$$L\left(u(x_i^{HF}), \hat{u}(x_i^{HF})\right) = \frac{\left\|u(x_i^{HF}) - \hat{u}(x_i^{HF})\right\|_p}{\left\|u(x_i^{HF})\right\|_p} \quad (8)$$

For a GCS problem aiming to predict pressure and $CO_2$ saturation, we observe that optimizing based upon the $l_2$-loss can achieve better model performance than based upon the $l_1$-loss, especially for CO2 saturation prediction. This finding is consistent with the experience reported by Wen et al. (2022). The learning rate of pre-training is recommended to be set larger than that of the fine-tuning to avoid large changes in model parameters (Lyu et al., 2023).

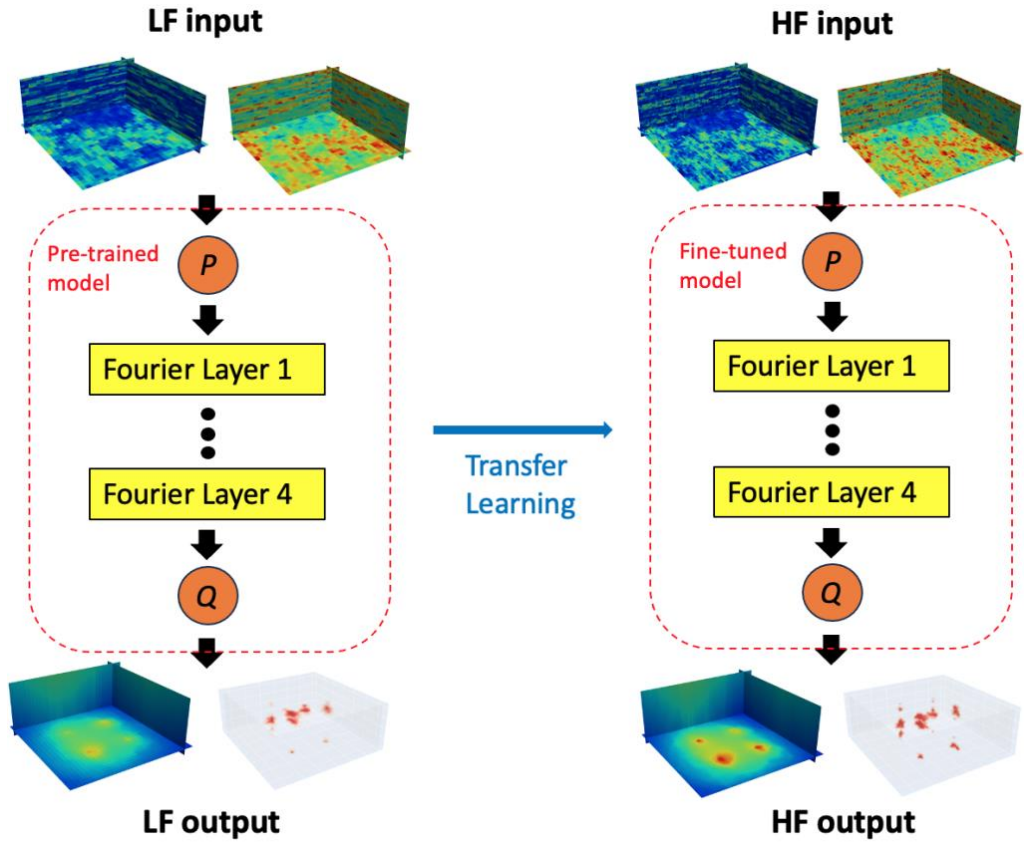

Figure 1. Multi-fidelity training framework based on FNO. HF stands for high-fidelity data and LF stands for low-fidelity data.

## 3. Problem setup and multi-fidelity model training details

We evaluate the efficacy of the multi-fidelity training framework through a synthetic large-scale GCS model. The model is generated based on the clastic shelf depositional environment (Bosshart et al., 2018) for dedicated $CO_2$ injection into deep saline aquifer. We summarize the details of the reservoir model and multi-fidelity models in the following sections.

3.1 Problem Setup

The 3D reservoir model has an aerial extent of 32,156m × 32,156m and thickness of 85m. The model is discretized into 64 × 64 × 28 grid cells in three dimensions. $CO_2$ is injected from four injection wells (thick red lines in Figure 2(a)) with a total injection rate of 2 million metric tons of $CO_2$ per year. The four injection wells are completed in all reservoir layers. A trained CNN-PCA model as described in H. Tang et al. (2022) is applied to generate 2800 geomodel realizations. The geomodel realizations are conditioned to facies type, porosity, and permeability at nine well locations (four injection wells and five exploration wells) shown in Figure 2(a). The 64 × 64 × 28 geomodel realizations are upscaled to 32 × 32 × 28 through Gaussian filtering and interpolation. We apply the open-source carbon storage simulator GEOS (Settgast et al., 2018) as the numerical reservoir simulator. The reservoir properties and models considered in the simulation are summarized in Table 1. The simulation results are output yearly for ten years. In Figure 2, we present a 64 × 64 × 28 and the correlated 32 × 32 × 28 realizations. The porosity and permeability fields of these two realizations are very similar with respect to features such as the correlation lengths being preserved. However, the pressure and saturation fields predicted by GEOS based on these two fields are different after 10 years of injection. The pressure magnitude of 32 × 32 × 28 realizations is lower than that of 64 × 64 × 28 realizations especially around the injection wells. The $CO_2$ saturation front of the 64 × 64 × 28 realizations is sharp, but that of the 32 × 32 × 28 realizations is diffused. The locations of $CO_2$ plumes are slightly different as well. These differences are mainly caused by resampling heterogeneous input fields between different grid resolutions and numerical errors associated with different grid resolutions, which exists for homogeneous fields as well. The differences can be mitigated by the multi-fidelity FNO model as shown in the following section.

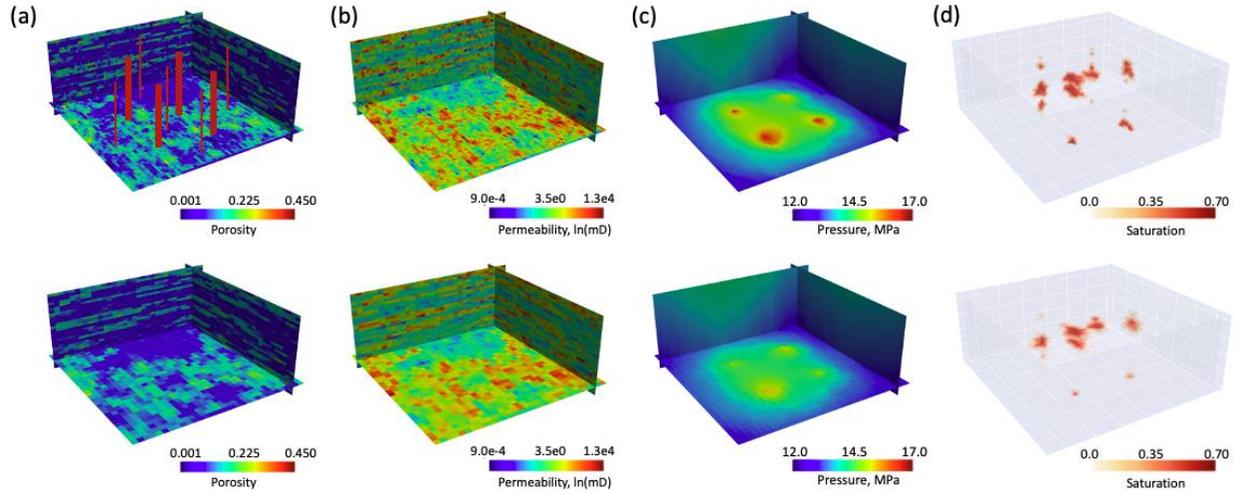

Figure 2. (a) Porosity, (b) permeability, (c) pressure, and (d) $CO_2$ saturation fields at ten years from a same case based on $64 \times 64 \times 28$ (top row) and $32 \times 32 \times 28$ (bottom row) discretizations. Pressure and saturation fields are obtained from GEOS simulations. The red lines in (a) indicate well locations. Thick lines represent injection wells and thin lines represent exploration wells.

Table 1. Reservoir properties and models applied in GEOS simulation.

| Parameters | Value |
| --- | --- |
| Grid resolution | $64 \times 64 \times 28$ or $32 \times 32 \times 28$ |
| Constitutive model | CO2-Brine Phillips fluid |
| Relative permeability | Brooks Corey relative permeability |
| Boundary Condition | Carter-Tracy analytical aquifer |
| Pore compressibility | $4.64 \times 10^{-9}$ Pa$^{-1}$ |

3.2 Training details of multi-fidelity FNO models

We use the $32 \times 32 \times 28$ realizations for our low-fidelity data and $64 \times 64 \times 28$ realizations for our high-fidelity data. For both datasets, we divide the 2800 realizations into 2500 for training, 200 for validation and 100 for testing. The FNO model considers four input features, porosity, logarithmic permeability, cumulative injection ratio, and time. The cumulative injection ratio is defined as the ratio of cumulative $CO_2$ volume being injected to each well. This input feature is essential for accurate $CO_2$ saturation prediction since it provides additional information about how the total $CO_2$ injection volume being distributed among each well. Porosity and logarithmic permeability are directly input as a 3D map after min-max normalization. Cumulative injection ratio and time are given as normalized values assigned to grid cells along injection well trajectories. Values of the rest of the 3D grids are set to be zero. The architecture of the FNO model and relevant output shapes for low-fidelity and high-fidelity data are summarized in Table 2. We apply a mode number of 20 for the x and y dimensions and a mode number of 10 for the z direction. A padding operator is applied to accommodate the non-periodic boundaries (Wen et al., 2023).

Table 2. FNO model architecture and output shapes for multi-fidelity data. *Cat* denotes concatenating with grids, *Spectral3d* denotes the operator in Fourier space defined in Eq. (5), *GELU* denotes Gaussian Error Linear Units layer.

| Layer | Operation | Output shape of low-fidelity data | Output shape of high-fidelity data |
| --- | --- | --- | --- |
| Input | - | (32, 32, 28, 4) | (64, 64, 28, 4) |
| Lifting ($P$) | Cat, Linear operator | (32, 32, 28, 16) | (64, 64, 28, 16) |
| Padding | Pad both sides with 6 | (44, 44, 40, 16) | (76, 76, 40, 16) |
| Fourier 1 | Spectral3d/Linear, Add, GELU | (44, 44, 40, 16) | (76, 76, 40, 16) |
| Fourier 2 | Spectral3d/Linear, Add, GELU | (44, 44, 40, 16) | (76, 76, 40, 16) |
| Fourier 3 | Spectral3d/Linear, Add, GELU | (44, 44, 40, 16) | (76, 76, 40, 16) |
| Fourier 4 | Spectral3d/Linear, Add, GELU | (44, 44, 40, 16) | (76, 76, 40, 16) |
| De-padding | De-pad both sides with 6 | (32, 32, 28, 16) | (64, 64, 28, 16) |
| Projection ($Q$) | Linear operator | (32, 32, 28, 1) | (64, 64, 28, 1) |

For the first step, we train a low-fidelity model based on the 2500 low-fidelity data set. The network is trained for 400 epochs using the adaptive moment estimation (ADAM) optimizer with a learning rate of 0.01 (Kingma and Ba, 2014). The best model based on the minimum $l_2$-loss of the validation dataset is saved as the final low-fidelity model (LF). For the second step, the pre-trained low-fidelity model is fine-tuned with high-fidelity data. The model is trained for an additional 200 epochs using the ADAM optimizer with a learning rate of 0.005. We randomly pick 100, 300, and 500 realizations in the 2500 training realizations for fine-tunning purpose. The trained multi-fidelity models are designated MF100, MF300, and MF500 respectively. For comparison purposes, we also train a high-fidelity FNO model (HF) based on the 2500 high-fidelity dataset.

We use Pytorch framework to build and train the multi-fidelity FNO models. Building on top of the original FNO structure, we implement our working pipeline on LLNL's Lassen computing clusters with 756 computing nodes, where each of the node contains 4 NVIDIA V100 GPUs. To take advantage of multi-node GPUs, we use Pytorch's DistributedDataParallel API for data distributed training across nodes. Since the full training dataset is too large to load into memory, we write a custom data loader to stream the training data to different GPU workers.

## 4. Performance evaluation of multi-fidelity models

### 4.1 Multi-fidelity FNO model performance

The model performance is evaluated based on the 100 hold-out high-fidelity testing data set. We apply the root mean square error (RMSE) as the metric for pressure prediction:

$$\text{RMSE} = \sqrt{\frac{1}{n_s}\frac{1}{n_t}\sum_{i=1}^{n_s}\sum_{t=1}^{n_t}\left\|\widehat{\boldsymbol{p}}_i^t - \boldsymbol{p}_i^t\right\|_2^2}, \qquad (9)$$

where $n_s$ is the number of testing realizations, $n_t$ is the number of testing time steps, and $\widehat{\boldsymbol{p}}_i^t$ is the FNO model predicted pressure for realization $i$ at timestep $t$. $\boldsymbol{p}_i^t$ is the pressure predicted by numerical simulators. For saturation prediction, we apply the plume mean absolute error (PME) as the testing metric (Wen et al., 2021):

$$\text{PME} = \frac{1}{\sum I_{t,i}} \frac{1}{n_s} \frac{1}{n_t} \sum_{i=1}^{n_s} \sum_{t=1}^{n_t} I_{t,i} |\widehat{S}_i^t - S_i^t|, I_{t,i} = 1 \; if \; S_i^t > 0 \tag{10}$$

where $\widehat{S}_i^t$ is the CO$_2$ saturation predicted by FNO models, and $S_i^t$ is the CO$_2$ saturation predicted by physical simulators. Since CO$_2$ saturation distributions contain a lot of zero values in the simulation domain, $I_{t,i}$ is only considered in regions where CO$_2$ plume exists.

The grid-invariant property of FNO allows the LF to be directly tested on the high-fidelity datasets. In Figure 3(a), we present the pressure RMSE of all high-fidelity testing cases as box plots for LF, MF100, MF300, MF500, and HF. Each box on the plot extends from the 25$^{th}$ to 75$^{th}$ percentiles of the pressure RMSE values. The solid orange line in the box indicates the median RMSE value. If we directly use LF to predict on the high-fidelity dataset, the mean RMSE value for all the testing cases is as large as 0.614 MPa. After being fine-tuned with the 100 high-fidelity data, the mean RMSE value drops significantly to 0.086 MPa. Increasing the number of training high-fidelity data to 300 and 500 only has minor impact on model performance. The mean RMSE value is 0.080 MPa for both the MF300 and MF500 models, which is very close to the performance of the high-fidelity model. As expected, the high-fidelity model has the lowest mean RMSE value, 0.072 MPa, on the testing dataset. In Figure 3(b), we plot the mean pressure RMSE evolution with time for MF100, MF300, MF500, and the HF models. The error bar indicates the range between 5$^{th}$ percentile and 95$^{th}$ percentile of the RMSE values at the specific timestep. The errors from the predicted pressure increase slightly with time for all the models. The error ranges of the MF models are slightly higher than that of HF models.

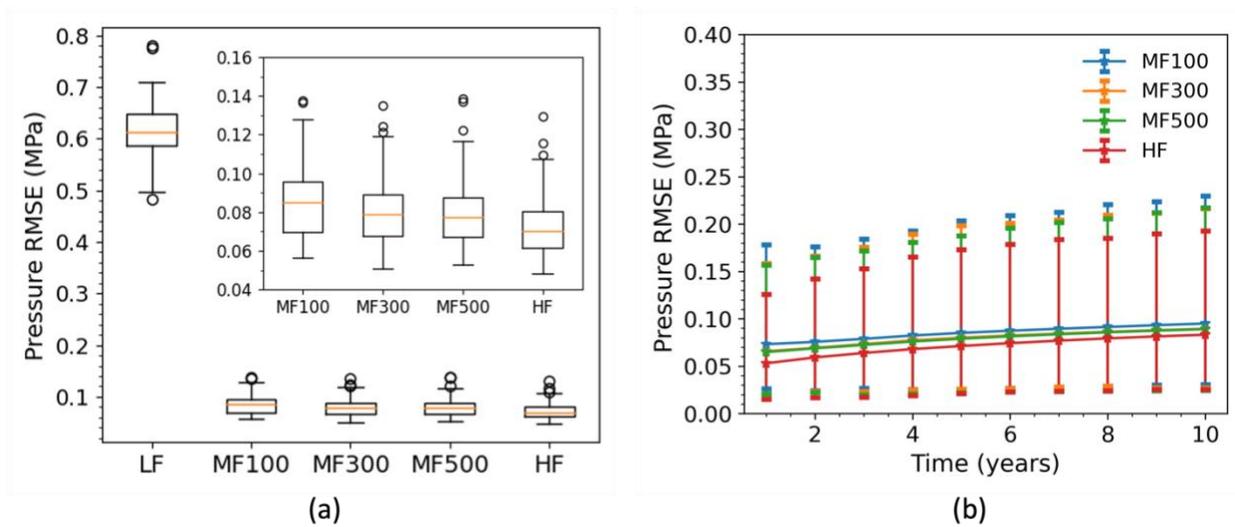

Figure 3. (a) Box plot of pressure root mean square errors (RMSE) for low-fidelity model (LF), multi-fidelity models trained with 100, 300, and 500 high-fidelity data (MF100, MF300, and MF500) and high-fidelity model (HF), the small insert picture shows the zoom-in view of the MF models results and (b) mean pressure RMSE with a 90% credible interval evolution with time for MF100, MF300, MF500, and HF based on 100 high-fidelity testing cases.

In Figure 4, we present pressure fields for three testing cases at the end of 10 years of injection. The first row displays pressure fields estimated by physics-based simulator (GEOS), the second row displays pressure fields predicted by HF, and the third row displays pressure fields predicted by MF100. Both HF and MF100 provide pressure predictions that are in close visual agreement with pressure fields from GEOS. To quantitatively show the difference, we also present the plots of pressure difference of these cases in Figure 5. The largest pressure error is around 1.2 MPa, appearing near the injection well locations. The difference plots also indicate that although the results from HF and MF100 are different, the prediction errors are within the same range for these two models.

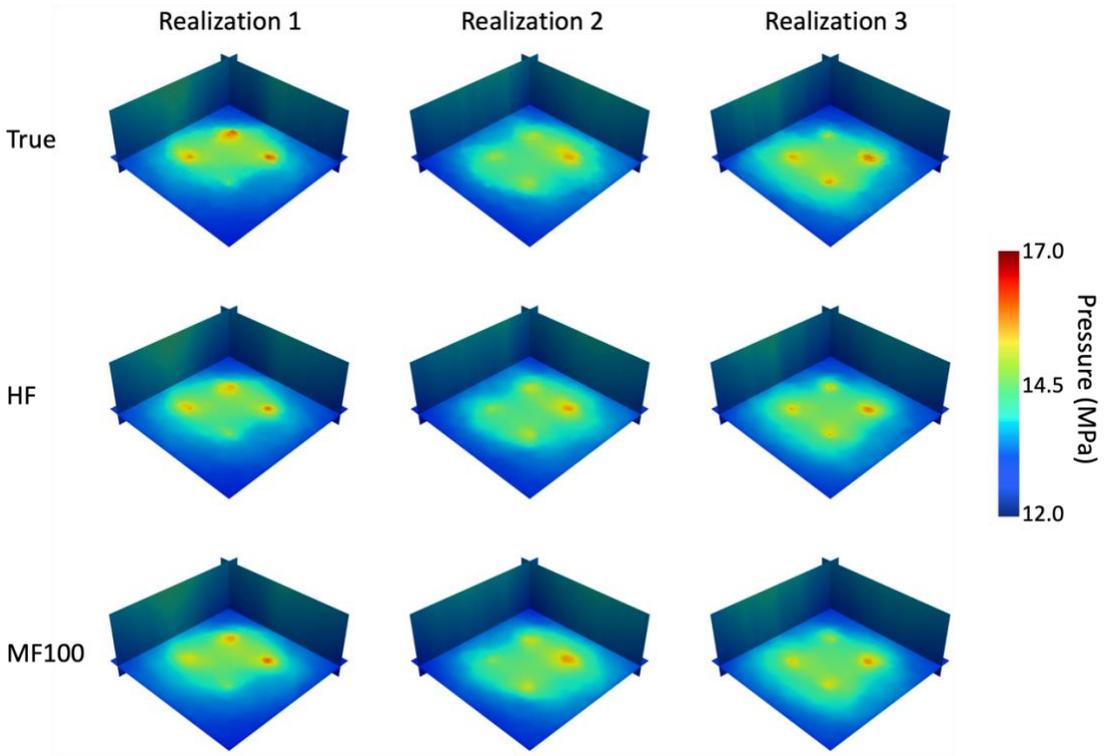

Figure 4. Comparison of true pressure fields predicted by GEOS simulations, MF100, and HF for three realizations from the high-fidelity testing dataset. All results are at 10 years.

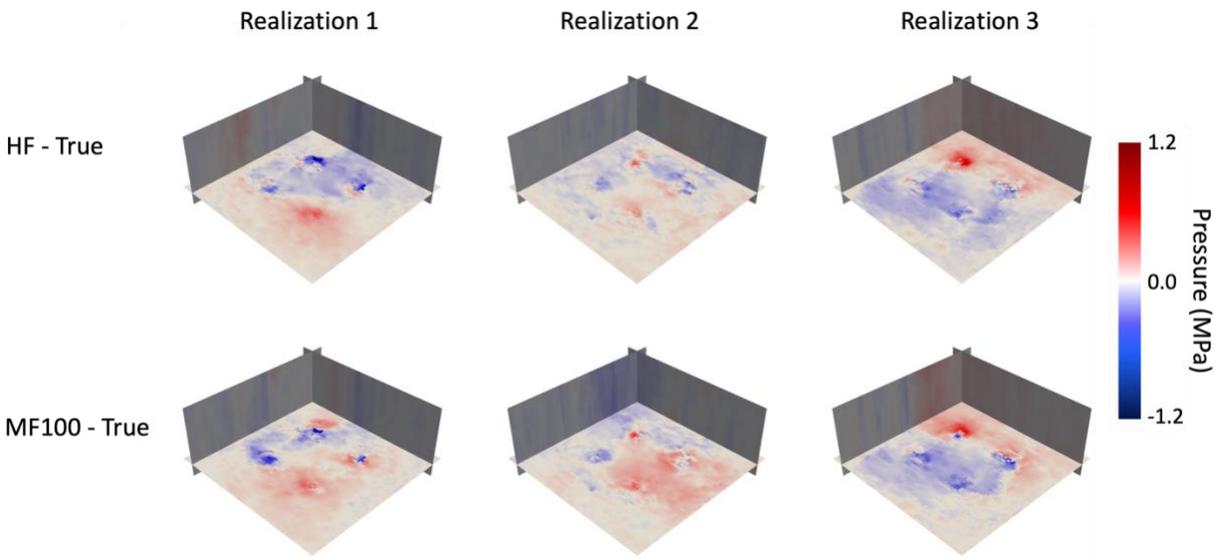

Figure 5. Comparison of pressure field differences of HF and MF100 minus true pressure fields from GEOS simulations. The three realizations are the same as that in Figure 4.

In Figure 6(a), we show the box plot of saturation PME for LF, three MF, and HF models. If directly using LF to predict the high-fidelity testing cases, the saturation prediction has an average PME error as large as 0.275. After being fine-tuned with the 100 high-fidelity training data, the average PME error of MF100 decreases to 0.145. The average PME errors of MF300 and MF500 are 0.139 and 0.134 respectively, which are very close to PME error of HF, 0.130. In Figure 6(b), we plot the mean saturation PME evolution with time for MF100, MF300, MF500 and HF models. The 90% credible intervals are also shown in the plot. The MF models have slightly wider credible intervals than the HF model does. The ranges of 90% credible intervals increase slightly via time for all models.

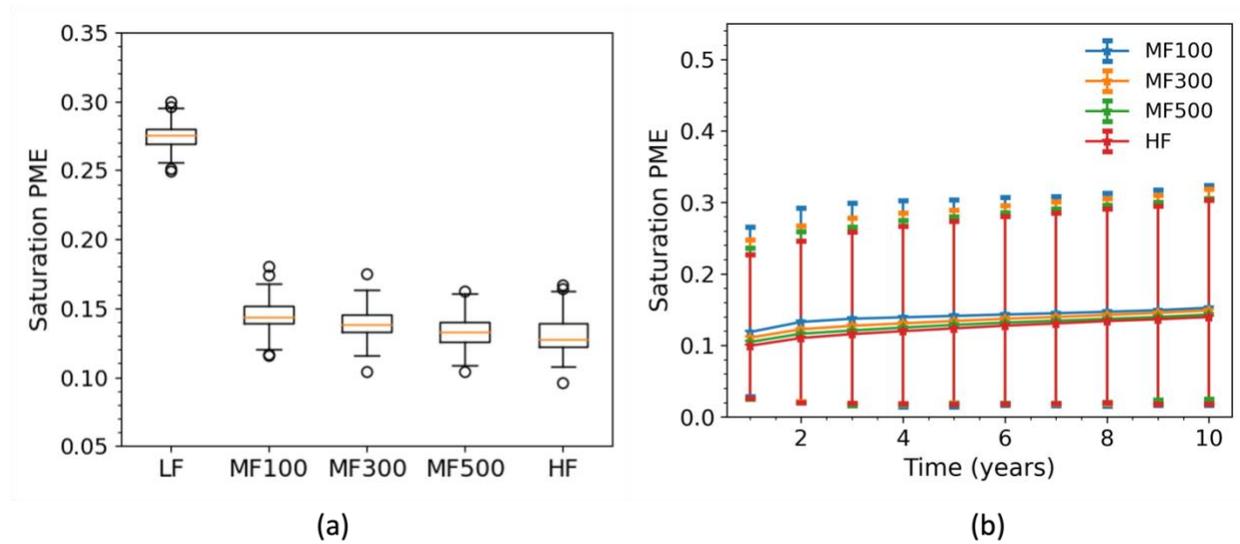

Figure 6. (a) Box plot of saturation plume mean absolute errors (PME) for LF, MF100, MF300, MF500, and HF and (b) mean saturation PME with a 90% credible interval evolution with time for MF100, MF300, MF500, and HF based on 100 high-fidelity testing cases.

In Figure 7, we present $CO_2$ saturation distributions after 10 years of injection for the same three cases as in Figure 4. The predictions from HF and MF100 are in close visual agreement with the GEOS predictions in terms of $CO_2$ plume shapes, sizes, and locations. We also make the plume footprint plots for $CO_2$ saturation as shown in the first row of Figure 8. We first apply a threshold of 0.1 to convert a 3D $CO_2$ saturation distribution to a binary field, then we project the binary field onto a 2D plane to obtain a plume footprint plot. Using plume footprint of $CO_2$ saturation prediction from GEOS as a reference plot, the plume footprint differences of HF and MF100 predictions are presented in the second and third rows of Figure 8. The plume footprint differences are comparable between these two models. As we observed for the pressure prediction results, fine-tuning MF models with 100 high-fidelity data is enough to achieve model performance as good as the corresponding HF model. We only observe minor improvement by using more high-fidelity data.

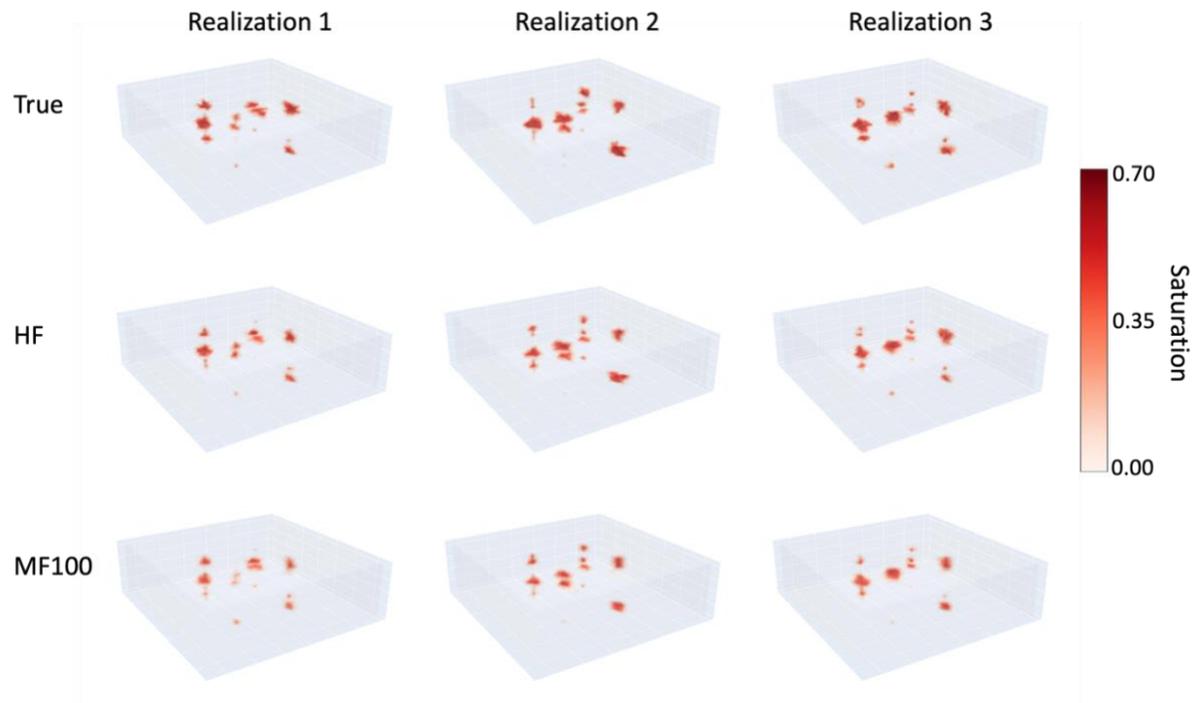

Figure 7. Comparison of true $CO_2$ saturation fields predicted by GEOS simulations, MF100, and HF for three realizations from the high-fidelity testing dataset. All results are recorded at 10 years.

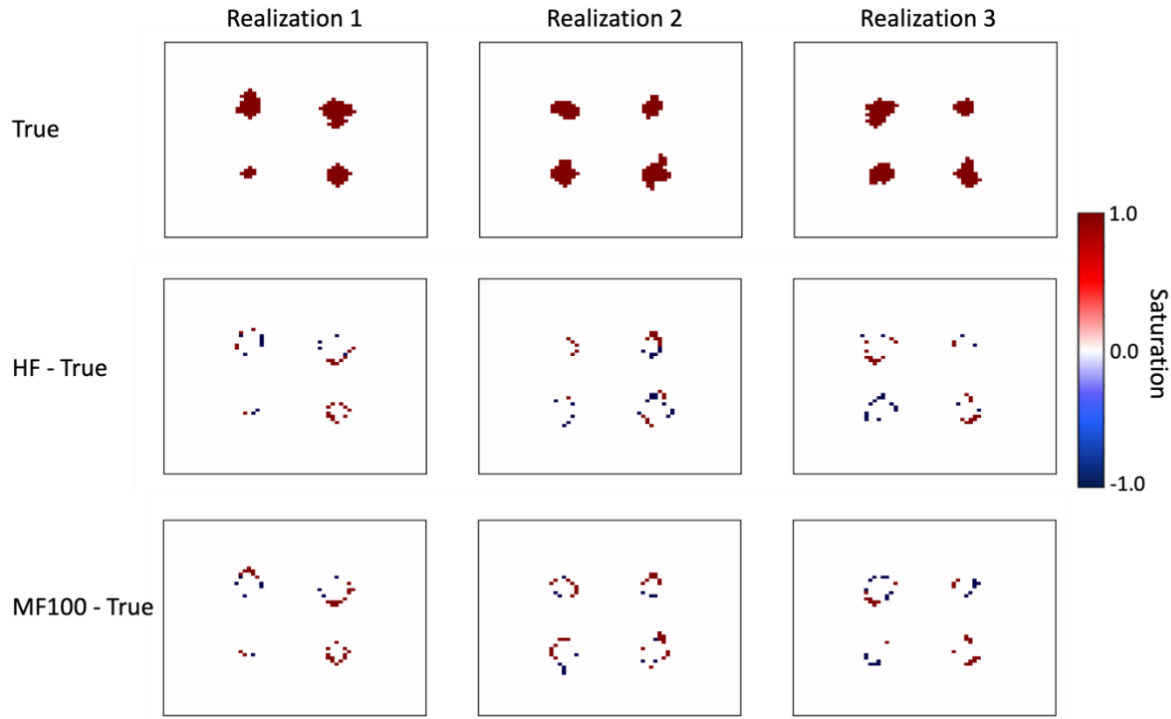

Figure 8. The first row shows the true plume footprint of $CO_2$ saturation simulated by GEOS for the same realizations shown in Figure 7. The second and third rows show differences of plume footprint predicted by HF and MF100 from that predicted by GEOS.

4.2 Computational cost comparison

It takes GEOS 2 CPU core hours to generate a high-fidelity ($64 \times 64 \times 28$) case and 0.3 core hour to generate a low-fidelity ($32 \times 32 \times 28$) case. Note that the grid cell count of the high-fidelity case is four times that of the low-fidelity case, while the computational cost of the high-fidelity case is almost seven times that of the low-fidelity case. GEOS applies a fully implicit time discretization, and we gradually increase the time step size to a fixed maximum value for each simulation. We observe that simulation on coarser grids converges faster (requiring less iterations) since there are less extreme porosity/permeability values. Another dominate factor of computational cost is linear solvers. In this case, we apply a generalized minimal residual (GMRES) iterative solver with a multigrid reduction (MGR) preconditioner.

The FNO models are trained on 8 Nvidia V100 GPUs, the average training costs per epoch for HF and MF100 are summarized in Table 3. The total training costs are calculated by adding up the training costs of first and second steps and then multiplying by the total number of GPUs. In the above case, the multi-fidelity training framework leads to an 81% reduction in computational cost associated with generating training data. In addition, the multi-fidelity training framework saves 45% of the time to train the pressure model and 35% of the time to train the saturation model.

As for GPU memory usage, all steps shared the same model structure, which consumes about 0.12GB of GPU memory. The total GPU memory required for training is mainly dominated by the dimension and batch size of input data. The advantage of having a data-paralleled 3D FNO workflow is that the model can be trained as long as the GPU has sufficient memory to process a single training sample. For an Nvidia V100 GPU with 16GB memory, the proposed 3D FNO workflow can train a 3D reservoir model with up to 2 million grids.

Table 3. Training times and computational costs of training datasets for HF and MF100 models.

|  | First step ($8 \times$ V100 GPUs) | Second step ($8 \times$ V100 GPUs) | Total training (GPU hours) | Data cost (CPU core hours) |
|---|---|---|---|---|
| HF/pressure | - | 76.68 s/epoch | 68.16 | 5400 |
| MF100/pressure | 38.14 s/epoch | 8.23 s/epoch | 37.56 | 1010 |
| HF/saturation | - | 52.35 s/epoch | 46.52 | 5400 |
| MF100/saturation | 30.46 s/epoch | 7.34 s/epoch | 30.34 | 1010 |

## 5. Model generalizability to indirectly correlated datasets

In the above case, the low-fidelity geomodels are directly upscaled from high-fidelity geomodels, and the neural network can learn most of the model behavior from the low-fidelity data and makes minor corrections utilizing a small amount of high-fidelity data. In this section, we are interested in exploring whether the multi-fidelity FNO model will work if the low-fidelity and high-fidelity datasets are generated differently, and thus not directly correlated, as well as limiting the total number of high-fidelity data to reduce computational cost.

We still focus on the same reservoir model as introduced in section 3.1, but with a finer discretization. The model is discretized with $211 \times 211 \times 28$ grid cells in three dimensions, resulting in approximately 1.2 million grid cells. Gaussian random function simulation (GRFS) is applied to generate heterogeneous porosity and permeability fields conditioned to the same facies distribution shown in Figure 9(a). Three typical sets of realizations at P25, P50, and P75 are selected by ranking them based on the ranges of porosity. As we consider realizations ranging from P25 to P75, the porosity distributions trend towards larger values. A P50 realization of porosity and permeability fields are depicted in Figure 9(b) and 9(c). The porosity and permeability distributions of P25, P50, and P75 realizations are shown in Figure 9(d) and 9(e). All the realizations were generated using Petrel by Energy & Environmental Research Center (EERC) of University of North Dakota under the U.S. DOE SMART Initiative (Yonkofski and McGuire, 2011). There are 22 P25 realizations, 23 P50 realizations, and 16 P75 realizations in total. EERC applied CMG-GEM (v2019) to run all the realizations for 10 years of $CO_2$ injection. $CO_2$ is injected via four injection wells with the same total injection rate of 2 million metric tons $CO_2$ per year and a maximum bottomhole pressure (BHP) constraint of 32.2MPa. All P50 and P75 realizations meet the total injection targets without reaching the BHP limit. Some P25 realizations reach the BHP limit and miss the total injection target. It took CMG-GEM about 37 CPU core hours to run a single realization with 10 years of injection.

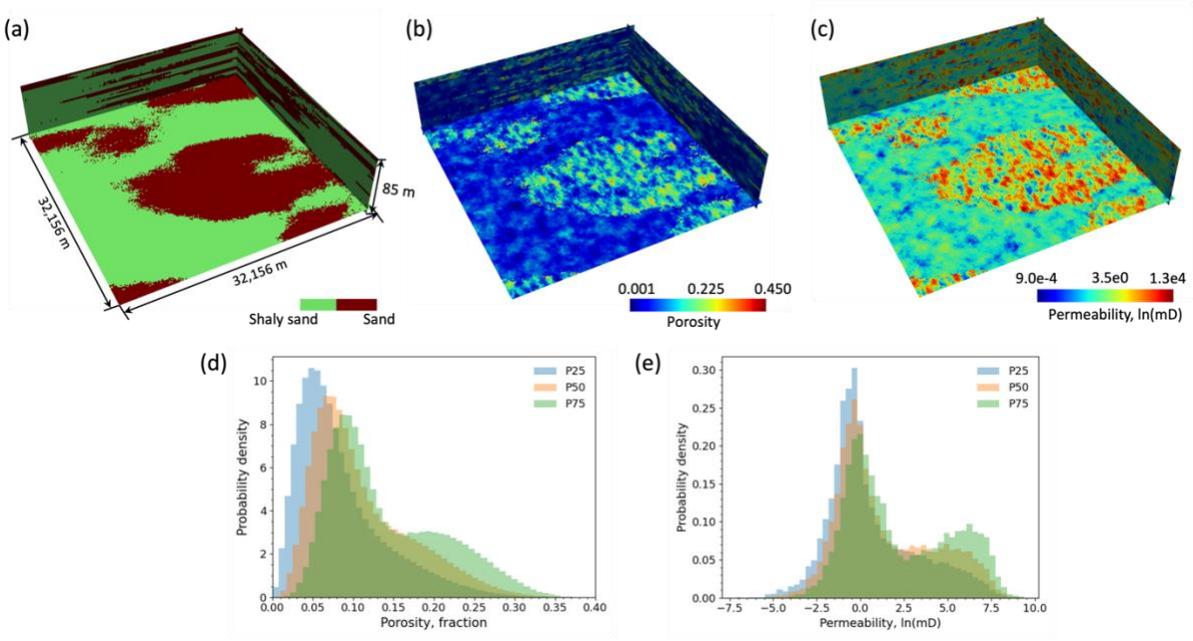

Figure 9. (a) Facies, (b) porosity, and (c) logarithmic permeability distributions from a P50 realization with a $211 \times 211 \times 28$ discretization. Histograms of (d) porosity and (e) logarithmic permeability for P25, P50, and P75 realizations.

All of the $64 \times 64 \times 28$ realizations described in section 3.1 are generated following the P50 model distribution. We explore the application of three kinds of low-fidelity models in this case: the trained HF model based on $64 \times 64 \times 28$ realizations (denoted as 64HF), the trained LF model based on $32 \times 32 \times 28$ realizations (denoted as 32LF), and the trained MF100 model based on the multi-fidelity dataset (denoted as 64MF). We utilize the $211 \times 211 \times 28$ realizations as high-fidelity data to fine-tune the model. We hold 2 P25 realizations, 2 P50 realizations, and 1 P75 realization for testing. The rest of the realizations are for training and validation. We observe that training pressure models separately based on P25, P50, and P75 realizations have better prediction accuracy than training a single model based on a mixed dataset. However, it is the opposite for the saturation model. Therefore, we train three separate pressure models with three sets of realizations and train a single saturation model based on all realizations. In Table 4, we summarize the average pressure RMSE and saturation PME values of all testing cases for multi-fidelity FNO models trained based on three low-fidelity models mentioned above. In Figures 10 and 11, we show the pressure and saturation fields predicted by different multi-fidelity FNO models (64HF, 64MF, and 32LF) for P25-1, P50-1, and P75-1 realizations and compare them with the results of same realizations simulated by CMG-GEM.

For pressure prediction, the multi-fidelity models trained based on 64HF and 64MF can predict pressure fields within a reasonable error range despite the extremely limited high-fidelity training data. The multi-fidelity model trained based on 32LF has much larger prediction errors than that trained based on 64HF and 64MF. It reveals the necessity to generate training datasets with multiple levels of fidelities if the target field has a high grid resolution (more than 1 million grids). For all three multi-fidelity models, the pressure errors for the P25 testing cases are significantly higher than that for the P50 and P75 testing cases. This is mainly because the pressure variability of the P25 realizations is significantly higher than that of other realizations due to the change of boundary conditions (changing from total injection rate constraint to constant pressure constraint).

Table 4. Testing pressure RMSE and saturation PME for multi-fidelity FNO models trained based on different low-fidelity models (64HF, 64MF and 32LF).

|  | P25-1 | P25-2 | P50-1 | P50-2 | P75-1 |
| --- | --- | --- | --- | --- | --- |
| Pressure RMSE (MPa)–64HF | 0.199 | 0.156 | 0.082 | 0.053 | 0.071 |
| Saturation PME–64HF | 0.206 | 0.210 | 0.202 | 0.203 | 0.203 |
| Pressure RMSE (MPa)–64MF | 0.215 | 0.160 | 0.102 | 0.068 | 0.087 |
| Saturation PME–64MF | 0.195 | 0.203 | 0.200 | 0.187 | 0.186 |
| Pressure RMSE (MPa)–32LF | 0.231 | 0.217 | 0.215 | 0.112 | 0.159 |
| Saturation PME–32LF | 0.203 | 0.213 | 0.216 | 0.189 | 0.188 |

For saturation prediction, the multi-fidelity models trained based on 64HF, 64MF, and 32LF have similar performance. The saturation prediction accuracy is similar for all testing cases as well. The multi-fidelity saturation models can roughly locate $CO_2$ plumes along the vertical wellbore, but the predictions for plume size and shape are not accurate. It is worth mentioning that we have more data (56 realizations) to fine-tune the low-fidelity saturation model compared to pressure models, but it is still not enough for the model to capture the detailed shapes of $CO_2$ plumes. We attempted to improve the saturation prediction through hyperparameter tuning such as increasing the mode truncation limit and changing the learning rate during fine tuning. We observed slight improvement by increasing the mode truncation to 40.

In this case, the low-fidelity and high-fidelity data are different from the following perspectives. First, the underlying geostatistical model of the two datasets are different: the low-fidelity geomodels are generated by SGS in GSLIB, but the high-fidelity geomodels are generated by GRFS in Petrel. Second, the porosity/permeability distributions of these two datasets are different: the high-fidelity data are following three different distributions (P25, P50, and P75), but the low-fidelity data are following a single P50 distribution. Third, the low-fidelity data are simulated by GEOS applying a simplified $CO_2$-brine model, and the high-fidelity data are simulated by CMG-GEM applying a full equation-of-state compositional simulation. All these differences make this multi-fidelity training task more challenging, yet informative. The two datasets are from different sources but describing the same reservoir with the same physical process. The results indicate that the proposed multi-fidelity FNO model can bridge between simulation results with slightly different physical assumptions, especially for pressure distributions. The multi-fidelity models can learn pressure diffusion better than $CO_2$ plume migration. It is mainly because that the saturation field is less continuous compared with pressure fields. The phase front of saturation field forms a contact discontinuity which is a challenging scenario for data-driven deep learning-based models to learn in general (Witte et al., 2023).

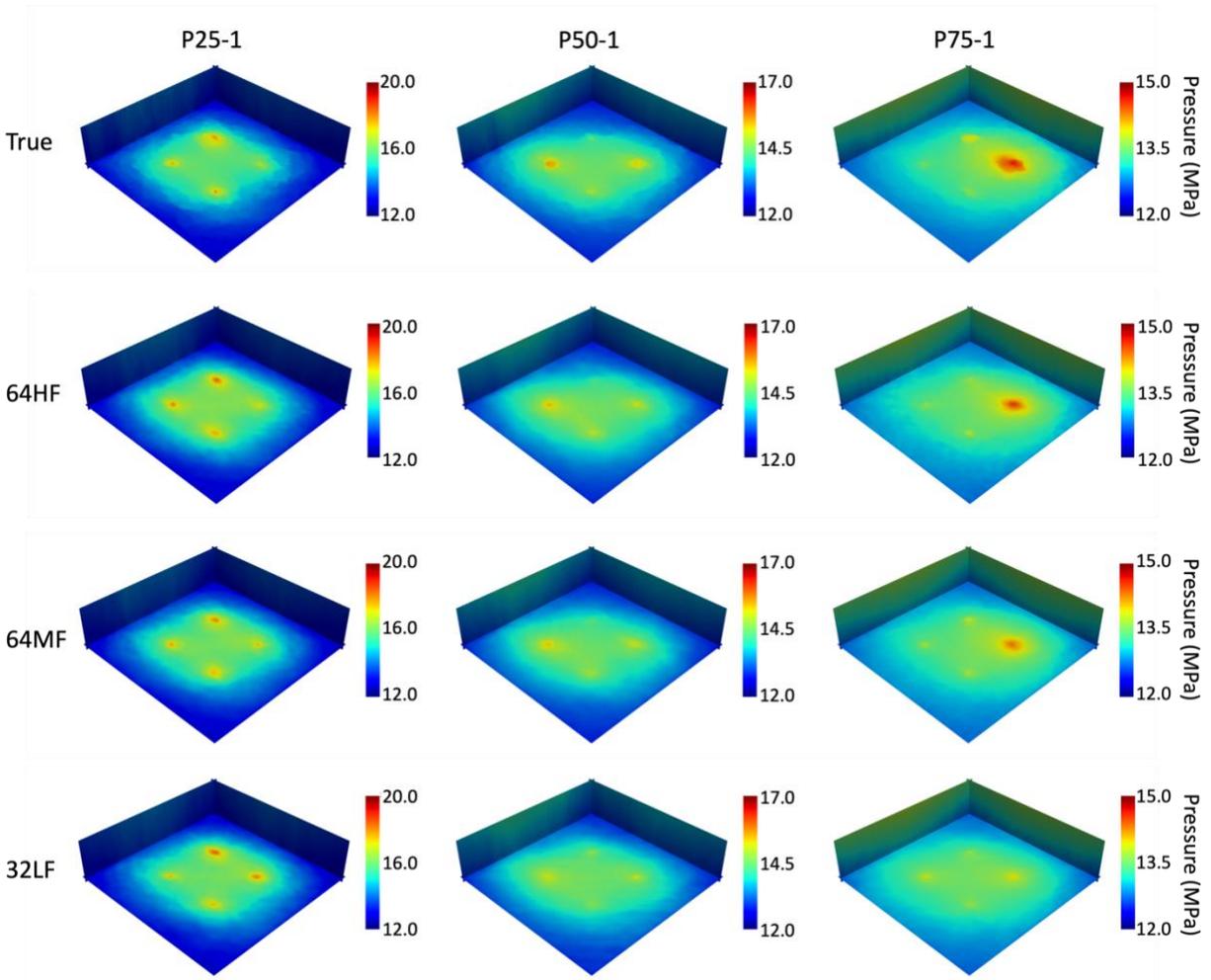

Figure 10. Comparison of true pressure fields predicted by CMG-GEM and multi-fidelity FNO models trained based on different low-fidelity models (64HF, 64MF, and 32LF) for P25-1, P50-1, and P75-1 realizations.

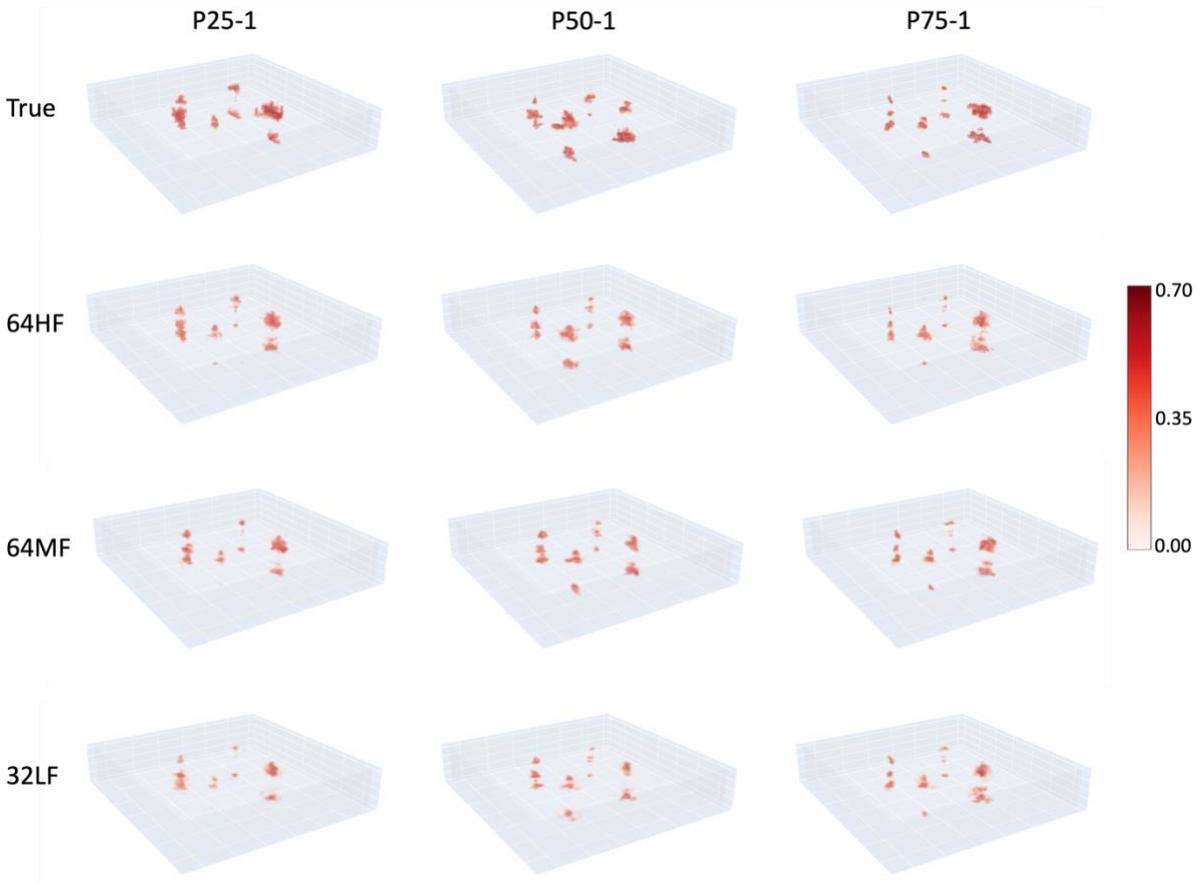

Figure 11. Comparison of true saturation fields predicted by CMG-GEM and multi-fidelity FNO models trained based on different low-fidelity models (64HF, 64MF, and 32LF) for P25-1, P50-1, and P75-1 realizations.

6. Discussion

The proposed multi-fidelity FNO model can be applied to build surrogate models for large 3D reservoir simulations with millions of grids. To build such a surrogate model is essential in speeding up a data assimilation workflow where model predictions are matched with monitoring pressure and $CO_2$ saturation data to reduce the uncertainty of prior porosity/permeability fields. The surrogate model can also be used as a quick forecasting tool for future reservoir pressure and $CO_2$ plume distribution after the porosity/permeability fields are being calibrated (H. Tang et al., 2022; Tang et al., 2021). In addition to pressure and $CO_2$ saturation distribution, $CO_2$ concentration in brine is also an important monitoring variable in GCS projects, which reflects the storage efficiency. If there are $CO_2$ concentration data available, the same multi-fidelity FNO model can be applied, and the trained surrogate model can help reduce the uncertainty of input geological models through a data assimilation workflow.

The multi-fidelity FNO model has the desirable grid-invariant property, which allows datasets with different grid resolutions sharing the same network structures and simplifies the transfer learning process. This advantage becomes more significant when we need to deal with data with more than two fidelity levels, as shown in the $211 \times 211 \times 28$ case. We previously applied a 3D recurrent R-U-Net structure to conduct transfer learning between the same $64 \times 64 \times 28$ and $32 \times 32 \times 28$ datasets (Jiang et al., 2021). We added an additional output layer to bridge between data with different fidelities. The transfer learning process has three steps: 1) training the base network with low-fidelity data, 2) adding a new output layer for high-fidelity data and training this layer with weights of other layers fixed, and 3) fine-tuning all the weights with high-fidelity data. The trained multi-fidelity R-U-Net model can achieve similar good performance for pressure prediction as shown in this paper. However, we cannot achieve similar good performance for saturation prediction with the same model structure.

The generalizability of deep neural network-based models is a known issue in scientific computation. In this study, we observe that the low-fidelity FNO model trained merely on low-fidelity data is not capable of making accurate predictions on high-fidelity testing data with 2x finer grids in x-y direction, although FNO-based architectures are supposed to learn an input-output mapping in the function space and to have 'super-resolution' properties (Li et al., 2021).

The discrepancy results from numerical errors and input field resampling errors between different grid resolutions. The state variables (i.e., pressure and saturation) computed based upon a coarse discretization are different from that computed based upon a fine discretization in terms of pressure magnitude and $CO_2$ plume shapes. Fine-tuning with high-fidelity data is always necessary to achieve desirable accuracy.

As a data-driven model, the multi-fidelity FNO model can only be applied to the same simulation domain with the same initial and boundary conditions. We test the model transferability between datasets with different porosity/permeability statistics and slightly different physical models caused by applying different reservoir simulators. The results indicate that the multi-fidelity FNO model is transferable among datasets describing the same reservoir model and operational conditions, especially for pressure prediction. On the other side, the results also reveal that the multi-fidelity FNO model can hardly give accurate predictions if the high-fidelity dataset has too many different patterns compared to the low-fidelity dataset. This issue becomes more significant in the prediction of $CO_2$ saturation.

## 7. Summary and Concluding Remarks

In this work, we propose a multi-fidelity FNO model to predict state variables (pressure and $CO_2$ saturation) in large-scale GCS problems. The multi-fidelity FNO model is established by first training a low-fidelity model with large quantities of low-cost, low-fidelity data and then fine tuning the model with a small amount of high-fidelity data. The grid-invariant property of the FNO model enables direct transfer learning between datasets using the same network structures. Some key findings are summarized as below.

We first test the efficacy of multi-fidelity FNO model for a case where the low-fidelity data (32 × 32 × 28) are directly upscaled from high-fidelity data (64 × 64 × 28). The multi-fidelity FNO model can provide results as accurate as the corresponding high-fidelity model by only fine-tuning with 100 high-fidelity data. The computational cost of generating training dataset for the multi-fidelity FNO model is reduced by 81%. The model training costs are reduced by 45% for the pressure model and 35% for the saturation model.

We further test the generalizability of multi-fidelity FNO model on the same reservoir model with a finer discretization (211 × 211 × 28). We observe that the multi-fidelity FNO model can predict pressure fields with acceptable accuracy after fine tuning with less than 20 high-fidelity data, even though the low-fidelity and high-fidelity data are generated from different geostatistical models and reservoir simulators. As for saturation prediction, the model can locate $CO_2$ plumes along wellbores, but cannot predict plume sizes and shapes accurately.

It remains an open question that whether deep learning-based models are scalable to 3D reservoir models with large size, complex geological features (e.g., faults), and complicated control strategies (e.g., time-dependent injection plans). For future work, we are exploring more generalizable models (such as physics-informed neural operators) to tackle the challenges associated with large-scale reservoir models and the practical limits of quantity and/or quality of training data.

## Acknowledgements

This manuscript has been authored by Lawrence Livermore National Security, LLC under Contract No. DE-AC52-07NA27344 with the US. Department of Energy (DOE). The United

States Government retains, and the publisher, by accepting the article for publication, acknowledges that the United States Government retains a non-exclusive, paid-up, irrevocable, world-wide license to publish or reproduce the published form of this manuscript, or allow others to do so, for United States Government purposes. The release number of the document is LLNL-JRNL-853052. This work was completed as part of the Science-informed Machine learning to Accelerate Real Time decision making for Carbon Storage (SMART-CS) Initiative (edx.netl.doe.gov/SMART). Support for this initiative came from the U.S. DOE Office of Fossil Energy's Carbon Storage Research program. Part of the implemented pipeline was supported by the LLNL-LDRD Program under Project No. 23-FS-021. The authors acknowledge Nicholas A. Azzolina and Matthew Burton-Kelly from Energy & Environmental Research Center, University of North Dakota for providing the training dataset. The codes of this paper are available at https://github.com/qingkaikong/3DFNO_GCS

# Nomenclature

*Notation*

| | |
|---|---|
| $a(x)$ | input space of neural operator |
| $D$ | computational domain |
| $g$ | gravity accelerating constant, $9.81 \text{m} \cdot \text{s}^{-2}$ |
| $k$ | absolute permeability |
| $k_r$ | relative permeability |
| $n_s$ | number of testing realizations |
| $n_t$ | number of testing timestep |
| $m$ | mass fraction |
| $p$ | pressure |
| $p_c$ | capillary pressure |
| $q$ | injected phase volume |
| $P$ | linear operator applied at the first layer of a neural operator |
| $Q$ | linear operator applied at the last layer of a neural operator |
| $R$ | learnable transformation in Fourier space |
| $S$ | saturation, m |
| $T_K$ | truncation operation to filter out Fourier modes higher than $K$ |

| *u(x)* | output space of neural operator |
| *v* | architecture of neural operator |
| **v** | phase transport velocity |
| *W* | linear operator |
| *x* | spatial discretization of domain *D* |
| *z* | depth |
| $\mathcal{F}$ | Fourier transform operator |
| $\mathcal{K}$ | kernel integral operator |
| *σ* | non-linear activation function |
| *ϕ* | porosity |
| *ρ* | density, kg·m$^{-3}$ |
| *μ* | viscosity, Pa·s |

*Subscripts*

| *c* | component |
| *l* | phase |
| *t* | layer number of neural operator |